\documentclass[sigconf]{acmart}
\usepackage{multirow}
\usepackage{makecell}
\usepackage{graphicx}
\usepackage{caption}
\usepackage{subfigure}
\AtBeginDocument{%
  \providecommand\BibTeX{{%
    \normalfont B\kern-0.5em{\scshape i\kern-0.25em b}\kern-0.8em\TeX}}}

\copyrightyear{2024}
\acmYear{2024}
\setcopyright{acmlicensed}\acmConference[SIGIR '24]{Proceedings of the 47th International ACM SIGIR Conference on Research and Development in Information Retrieval}{July 14--18, 2024}{Washington, DC, USA}
\acmBooktitle{Proceedings of the 47th International ACM SIGIR Conference on Research and Development in Information Retrieval (SIGIR '24), July 14--18, 2024, Washington, DC, USA}
\acmDOI{10.1145/3626772.3657671}
\acmISBN{979-8-4007-0431-4/24/07}

\begin{document}

\title{An Integrated Data Processing Framework for Pretraining Foundation Models}

\author{Yiding Sun\textsuperscript{$\dagger$} }
\affiliation{%
  \institution{Gaoling School of Artificial Intelligence, Renmin University of China}
  \city{Beijing}
  \country{China}
}
\email{emanual20.sun@gmail.com}

\author{Feng Wang\textsuperscript{$\dagger$} }
\affiliation{%
  \institution{Gaoling School of Artificial Intelligence, Renmin University of China}
  \city{Beijing}
  \country{China}
}
\email{rucwangfeng@gmail.com}

\author{Yutao Zhu}
\affiliation{%
  \institution{Gaoling School of Artificial Intelligence, Renmin University of China}
  \city{Beijing}
  \country{China}
}
\email{yutaozhu94@gmail.com}

\author{Wayne Xin Zhao}
\affiliation{%
  \institution{Gaoling School of Artificial Intelligence, Renmin University of China}
  \city{Beijing}
  \country{China}
}
\email{batmanfly@gmail.com}

\author{Jiaxin Mao\textsuperscript{*}}
\affiliation{%
  \institution{Gaoling School of Artificial Intelligence, Renmin University of China}
  \city{Beijing}
  \country{China}
}
\email{maojiaxin@gmail.com}

\renewcommand{\shortauthors}{Yiding Sun and Feng Wang, et al.}

\begin{abstract}
The ability of the foundation models heavily relies on large-scale, diverse, and high-quality pretraining data. 
In order to improve data quality, researchers and practitioners often have to manually curate datasets from difference sources and develop dedicated data cleansing pipeline for each data repository.    
Lacking a unified data processing framework, this process is repetitive and cumbersome.
To mitigate this issue, we propose a data processing framework that integrates a Processing Module which consists of a series of operators at different granularity levels, and an Analyzing Module which supports probing and evaluation of the refined data. 
The proposed framework is easy to use and highly flexible.
In this demo paper, we first introduce how to use this framework with some example use cases and then demonstrate its effectiveness in improving the data quality with an automated evaluation with ChatGPT and an end-to-end evaluation in pretraining the GPT-2 model. 
The code and demonstration video are accessible on GitHub\footnote{https://github.com/Emanual20/Yulan-GARDEN}.
\end{abstract}




\maketitle

\let\thefootnote\relax\footnotetext{$\dagger$ Equal contribution.}
\let\thefootnote\relax\footnotetext{* Corresponding author.}

\section{Introduction}



The exceptional performance of large language models~(LLMs) in numerous downstream tasks stems from the extensive knowledge of the foundation models~\cite{T5,llama2,glm130b,survey-llm4recsys,survey-llm4agents,survey-llm4ir}. 
According to the scaling law~\cite{scaling-law}, the performance of these foundation models depends on the large-scale, high-quality preprocessed pretraining data. 
The inclusion of diverse large-scale pretraining data aids in enhancing the generalization ability~\cite{bloom,palm,gpt3}, while high-quality data leads to improved training efficiency and effectiveness of LLMs~\cite{phi-1.5,llm-survey-zhaoxin,survey-llm-data-management}.

However, preprocessing the pretraining data is a challenging and time-consuming task. 
First, data comes from various sources~\cite{the-pile,bigscience-roots,redpajama-data,WuDao}, and different LLMs rely on different data recipes~(\textit{i.e.}, data mixture proportions)~\cite{llama,bloom,palm,chinchilla,DoReMi}, each with its corresponding processing pipeline~\cite{ccnet}. 
Reproducing such a process for each data repository becomes complicated and time-consuming. 
Additionally, there is a lack of standardized evaluation metrics for pretraining data. 
So the practitioners often have to use the performance in downstream tasks to measure the quality of the pre-training data. 
This further complicates the data collection and reprocessing for foundations models as training and fine-tuning these models need high computational costs, especially considering the current trend of exponential growth in model parameters~\cite{llm-survey-zhaoxin}.

\begin{figure*}[ht]
  \centering
  \includegraphics[width=0.9\linewidth]{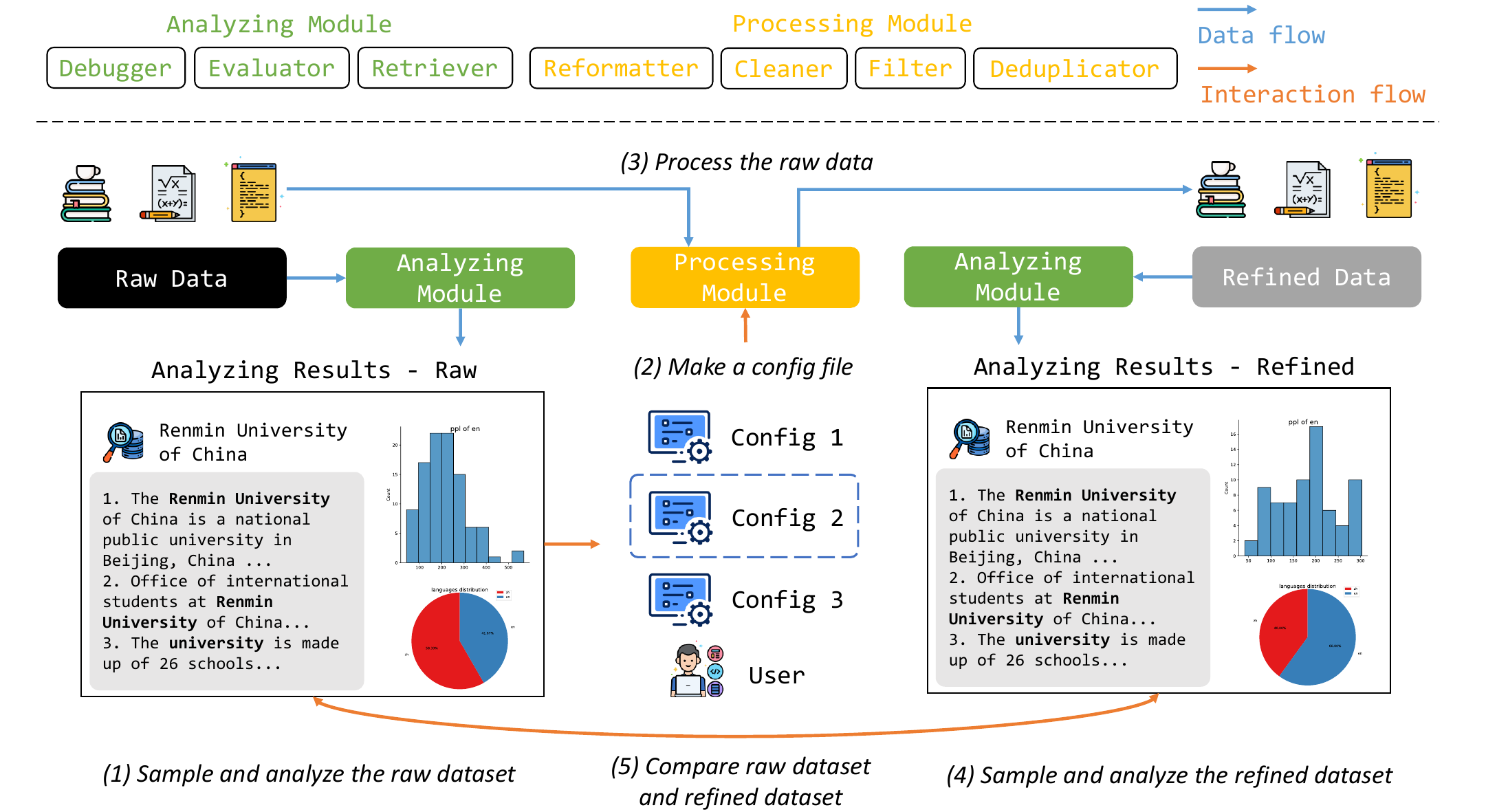}
  \setlength{\abovecaptionskip}{0.1cm}
  \setlength{\belowcaptionskip}{-0.6cm}
  \caption{Overview of our data processing framework.}
  \label{fig:framework}
\end{figure*}

To address the aforementioned issues, we propose an integrated data processing framework for pretraining data of foundation models. 
This framework consists of two main modules: \textbf{Processing Module} and \textbf{Analyzing Module}, as illustration in Figure~\ref{fig:framework}. 
The Processing Module comprises data processing operators of varying granularity, primarily implemented by heuristic and language model-based rules. 
These operators preprocess data from granularities of document, paragraph, and sentence. 
The Analyzing Module consists of Evaluator, Retriever, and Debugger, which can assist users in gaining an intuitive understanding of the dataset. 
Therefore with this framework, users can first probe the data to obtain preliminary insights with the Retriever and then customize data cleansing pipelines by flexibily combining and scheduling predefined operators provided in the Processing module, without coding manually.
Additionally, users can utilize the Evaluator in Analyzing Module to reconfirm whether the refined dataset meets training requirements, enabling further iterations to enhance data quality.

We further conduct two experiments to validate the effectiveness of the proposed framework in improving the data quality. 
Through automated evaluation with ChatGPT, we observe a significant improvement in the quality of refined dataset when applied to OpenWebText2, Wikipedia, and HackerNews~\cite{the-pile}. 
In the end-to-end evaluation, we train two GPT-2 models using CommonCrawl before and after processing, respectively. 
The model trained on the refined data demonstrates remarkable performance enhancement across downstream tasks of language modeling compared to the baseline.


\section{USE CASES}


\subsection{Integrated Data Processing Framework}


The inherent characteristics of datasets essentially influence the behavior of machine learning models~\cite{gebru2021datasheets}.
Consequently, as illustrated in Figure~\ref{fig:framework}, before the data-refined processing pipeline, users can probe into the raw dataset to gain insights into the dataset distribution, retrieve specific texts, and determine the parameters of the configuration file by Analyzing Module. 
The Evaluator demonstrates the distribution of text length, perplexity, language, etc. 
While the Retriever handles users' search queries and retrieve documents, the Debugger presents matching cases of Cleaner and the relation between filter parameters and the filter ratio of Filter.


After fetching results from the Analyzing Module, users can customize configuration files to meet dedicated data processing needs, e.g.~filter short texts, remove exact strings, and deduplicate. 
With the guidance of a user-defined configuration file, the submodules of the Processing Module collaborate to refine the raw dataset. 

Finally, users can probe into the refined dataset by Analyzing Module to compare the difference with the raw dataset. 
If the refined dataset still fails to fulfil the criteria of pretraining foundation models, users can modify the configuration file based on existing deficiencies and repeat the above steps until the data meets users' requirements. 
For example, users can remove gibberish fragments without actual symantic by setting the parameter `fil\_ppl' as 3 to filter texts with perplexity over avg.3$+s\cdot$std. 
If users still find gibberish in the refined dataset, they can adjust the parameters and reprocess the data until it meets the requirements for training.


\subsection{Retriever Augmentation}
In the pretraining stage, foundation models may suffer from hallucination, due to the inconsistent distribution of entities between pretraining data and users' real needs. 
Existing work introduced Retrieval Augmented Generation~(RAG) to involve external knowledge to mitigate hallucination~\cite{RAG-hallucination-benchmark-AAAI2024}.

In our framework, the Retriever Module can help researchers detect whether specific entity-level or topic-level texts exist in pretraining datasets by retrieving relevant keywords.
For example, when users' language model checkpoint has no knowledge about ``Renmin University'', to bridge the gap between the facts and the model’s generation results, users can search information about Renmin University in the dataset by Retriever. 
If there is no relevant information about Renmin University in retrieved texts, users will need to gather relevant texts from other datasets by Retriever to get a specialized dataset of Renmin University.

\section{Implementation}

\subsection{Processing Module}

Processing Module can refine the raw datasets by unifying different formats, filtering and denosing irrelevant information, and deduplicating. The Processing Module consists of four components:~Reformatter, Filter, Cleaner, and Deduplicator.

\noindent\textbf{Reformatter.}\quad
To accommodate datasets from diverse sources, we reformat the raw dataset into jsonlines, where each row is a JSON dictionary including a ``text'' field and other meta information, aligning with the format of Huggingface datasets.

\noindent\textbf{Filter.}\quad
Filter fixes the datasets by discarding contentless texts (e.g.,~crawled thin content webpage), non-target language texts, ambiguous texts, and toxic texts.
Filter excludes the texts from two perspectives. 
First, it employs labels annotated by trained language models, such as language and perplexity. 
Second, Filter excludes texts according to some handcrafted features, such as the occurrence of dirty words, the proportion of Alphabet or Chinese characters, the proportion of short lines, and number of entities.

\noindent\textbf{Cleaner.}\quad
Due to the specific formatting of data sources, some datasets may include content with weak semantics or with personal privacy, such as HTML tags in crawled web data or reporter names in news data.
Discarding such data directly would significantly reduce the scale of the training corpus, especially considering some rare data sources.
Consequently, we design Cleaner for eliminating useless information while keeping useful texts.
Cleaner fixes texts on the dimensions of strings, lines, and paragraphs using exact match and regular expression match. 
Additionally, Cleaner can convert the texts into another target language 
and extract contents from files of specific formats, such as HTML, epub, mobi, and PDF.

\noindent\textbf{Deduplicator.}\quad
The repetition in a dataset can potentially cause the strong double descent phenomenon~\cite{nakkiran2021deep}, indicating repetition may lead to an increase in test loss midway through training. 
This phenomenon highlights the importance of discarding repeated data before the training process to prevent such degradation in performance.
Therefore, Deduplicator utilizes the MinHash Locality Sensitive Hashing~(MinHashLSH)~\cite{indyk1998approximate, kocetkov2023the,chenghao_mou_2023_8364980} algorithm to remove duplicate texts from datasets. 
We split the texts into n-gram sequences, where n is set as 10. The similarity threshold is 0.7.

\subsection{Analyzing Module}

Analyzing Module helps users facilitate a more profound comprehension of datasets through statistics analysis, specific domain knowledge retrieval, and parameter analysis of Filter and match cases of Cleaner.
There are three components in the Analyzing Module, including Evaluator, Retriever, and Debugger.

\noindent\textbf{Evaluator.}\quad
To conduct the analysis of statistical features of both the raw dataset and refined dataset, Evaluator employs a visualization approach to present the data distribution (text length, perplexity, language, etc.) in a user-friendly manner.

\noindent\textbf{Retriever.}\quad
During the training process, checkpoints of foundation models may face the issue of being clueless about an entity, leading to the need to supplement specific entity or topic-level knowledge.
Therefore, Retriever integrates a simple search engine implemented by ElasticSearch to retrieve entity-level or topic-level texts, which can help users determine whether the dataset contains texts related to the field. 
The similarity of text is modeled by BM25. 
The index is divided into 20 shards for distributed storage and processing across multiple nodes. 

\noindent\textbf{Debugger.}\quad
Many parameters in the Filter and Cleaner cannot be directly determined before the refining process.
Consequently, we design Debugger to demonstrate the filtering or cleansing effects induced by different parameters.
To help users confirm the parameters of Filter and Cleaner, it samples the raw dataset, provides the relation between the parameters of Filter operators and the filtering ratio, and presents the match cases of Cleaner operators.

\section{Experiments}

We further evaluate the effectiveness of our data processing framework.
Specifically, we test whether there is an enhancement in data quality and whether the enhancement contributes to improved performance of the training process of foundation models.
Therefore, we conduct automated ChatGPT pairwise comparisons and end-to-end model training experiments to answer aforementioned research questions.

\subsection{Automated Evaluation}

\noindent\textbf{Implementation Details.}\quad
We adopt ChatGPT as an automated evaluation tool for data quality, utilizing its powerful ability of instruction following. 
We evaluate which data~(\textit{i.e.}, before and after processing) is more suitable for training LLMs. 
Specifically, ChatGPT is prompted to consider various dimensions such as format, fluency, coherence, and informativeness, and finally make pairwise judgments for which text is more suitable for pre-training a foundation model.
Data pairs that are identical before and after processing, or exceed the ChatGPT's context length limitation are excluded. 

\noindent\textbf{Datasets.}\quad
We focus on three distinct subsets~(Openwebtext2, Wikipedia-en, and HackerNews) of The Pile, representing different data sources~(web pages, encyclopedias, and news).
The details of data processing recipes can be found in our Github repository.

\noindent\textbf{Results.}\quad
The experimental results are presented in Table~\ref{tab:pairwise-evaluation}.
On all datasets, our processed data significantly outperforms the raw dataset.
Through case studies, We observe three factors that may lead to negative evaluation results. 
One is webpage tail information, such as "Read more on other websites" in Openwebtext2 and "References" in Wikipedia. 
Another one is paragraphs that are unrelated to the central semantics, such as advertisement. 
The third is useless characters, which may related to the webpage format.

\begin{table}[t]
  \setlength{\abovecaptionskip}{0.2cm}
  \setlength{\belowcaptionskip}{0cm}
  \begin{center}
    \caption{Results of Automated Evaluation. ChatGPT is prompted to compare the data quality of 500 data pairs before and after processing, reporting the number of win/lose/tied cases as metrics. ``Win'' indicates better quality than raw data.}
    \label{tab:pairwise-evaluation}
    \begin{tabular}{lccc}
      \toprule
       Datasets & \#Win & \#Lose & \#Tied \\ \midrule
      Openwebtext2 & 338 & 162 & 0\\ 
      Wikipedia(en) & 333 & 161 & 6\\ 
      HackerNews & 382 & 112 & 6\\ 
      \bottomrule
    \end{tabular}
  \end{center}
  \vspace{-0.5cm}
\end{table}

\subsection{End-to-end Evaluation}

To evaluate the data quality more intuitively, we trained a GPT-2 model using the raw and refined data respectively, denoted as GPT-2-raw and GPT-2-ref. 
We evaluated the language modeling capabilities of LLMs on other corpora in an end-to-end manner.

\noindent\textbf{Implementation Details.}\quad
Due to the limitation of computation resources, we select the 110M GPT-2~\cite{gpt2} as our foundation model. 
As the pretraining data WebText for GPT-2 is not fully open-source, we opt for a substitute pretraining dataset CommonCrawl, which is in the same category~(\textit{i.e.}, web pages) as WebText. 
To train the language model, we randomly initialize the parameters and used the raw and refined versions of CommonCrawl as the training data respectively. 
We train 12B tokens with a learning rate of 5e-4 and context length of 1024, utilizing one RTX 3090 GPU. 
To speed up the training process, we adopt FP16 mixed-precision training.

\noindent\textbf{Evaluation Metrics and Datasets.}\quad
To align with GPT-2, we employ datasets mentioned in the research paper to evaluate the language modeling capabilities~\cite{gpt2}.
These datasets include LAMBADA~\cite{lambada}, CBT-CN, CBT-NE~\cite{CBT}, WikiText103~\cite{wikitext103}, and 1BW~\cite{1BW}. 
These datasets have a relatively low overlap rate with CommonCrawl~\cite{gpt2}, which allows for evaluating the foundation model's ability to acquire language modeling skills and knowledge from pretraining data effectively. 
CBT-CN and CBT-NE are evaluated by Accuracy~(ACC), while the others are evaluated by Perplexity~(PPL).

\noindent\textbf{Results.}\quad
The evaluation results are shown in Table~\ref{tab:language_modeling} and Figure~\ref{fig:evaluation}.
Notably, GPT-2-ref achieves the same loss as GPT-2-raw after only 0.25M steps, whereas GPT-2-raw reaches that level after 2M steps. 
Across all datasets, GPT-2-ref exhibits superior performance compared to GPT-2-raw. 
Additionally, the PPL of GPT-2-ref on the LAMBADA and WikiText103 datasets demonstrate a noticeable trend of faster decrease compared to GPT-2-raw in the initial stage.
It indicates that utilizing our proposed data processing framework enhances both the efficiency and effectiveness of training foundation models.
However, as the training progress, the rate of PPL reduction for GPT-2-ref slows down, ultimately resulting in only a marginal lead over GPT-2-raw. 
We speculate that this is due to repeating the refined data for more than four epochs when training GPT-2-raw. 
We manage to ensure consistency with the number of tokens used in training GPT-2-raw, given the limited size of the refined CommonCrawl dataset.
This finding also poses a requirement for future works, whereby improving data quality needs to be balanced with the retention of an adequate quantity of data, to avoid loss oscillations caused by the repetition of training corpus.


\begin{table}
  \setlength{\abovecaptionskip}{0.2cm}
  \setlength{\belowcaptionskip}{0cm}
  \caption{Capability of Language Modeling.}
  \label{tab:language_modeling}
  \begin{tabular}{lccccc}
    \toprule
     &\thead{LAMBADA\\(PPL)} & \thead{WikiText\\(PPL)} & \thead{1BW\\(PPL)} &\thead{CBT-CN\\(ACC)} & \thead{CBT-NE\\(ACC)} \\
    \midrule
    GPT-2-raw& 134.04& 97.32& 220.50 & 61.05& 44.48\\
    GPT-2-ref& \textbf{122.43}& \textbf{81.98}& \textbf{175.59} & \textbf{72.60}& \textbf{50.98}\\
    \bottomrule
  \end{tabular}
  \vspace{-0.5cm}
\end{table}


\section{Related Works}

\noindent\textbf{Data Quality Improvement.}\quad
Current researches make efforts to improve pretraining data quality through deduplication, quality filtering, and increasing diversity~\cite{survey-llm-data-management}. 
\citet{dedup-acl2022} indicates removing duplicate data allows models to achieve comparable or better performance with fewer training steps. 
Regarding quality filtering, existing works rely on handcrafted heuristics rules or constructing neural classifiers to retain high-quality data~\cite{gpt3,palm,glam}. 
However, the limited efficiency of neural methods hinders their application to massive datasets. 
Furthermore, \citet{phi-1} and \citet{phi-1.5} select textbook-quality data from web pages to train a 1.3B model, surpassing larger LLMs in commonsense and code tasks.
While these works have shown individual improvements, there is still a lack of research considering data quality enhancement from a global perspective. 
This calls for attention from the community towards a unified data processing framework.

\noindent\textbf{Data Processing System.}\quad
Numerous open-sourced datasets are accompanied by their own processing pipeline. 
For instance, \citet{ccnet} propose CCNet to filter high-quality web pages from CommonCrawl dataset. 
As large-scale pretraining corpus from diverse sources, BLOOM~\cite{bloom} and RedPajama~\cite{redpajama-data} also provide details of their data construction process.
While, the process are not universally applicable and can only be employed in reproducing specific dataset.
The exploration of unified data processing frameworks remains scarce. 
As concurrent works, \citet{oasis} presents an interactive data management and evaluation system Oasis, whereas \citet{data-juicer} similarly proposes a data processing framework data-juicer comprising multiple operators. 
In comparison to our paper, Oasis aims to provide a more comprehensive assessment, while its curation module is limited to filtering without cleaning capabilities.
Additionally, data-juicer focuses on optimizing the fusion and rearrangement of multiple operators, significantly enhancing processing efficiency and ensuring data quality. 
However, our work includes additional practical and valuable functionalities, such as offering retrieval augmentation during the pretraining stage to supplement entity or topic-level knowledge for foundation models.

\section{Conclusion}

In this paper, we propose an integrated data processing framework. 
Users can customize the configuration of large-scale data to operate multi-level operators, enhancing data quality without coding manually. 
Additionally, users can employ the retrieval and evaluation modules to mitigate the issue of hallucinations during the pretraining stage. 
Our framework demonstrates improved data quality evaluated by ChatGPT on a subset of The Pile. 
Through an end-to-end evaluation approach, the foundation model trained on processed data exhibits superior generalization ability and language modeling capacity.


\begin{figure}[t]
  \centering  
  \includegraphics[width=0.8\linewidth]{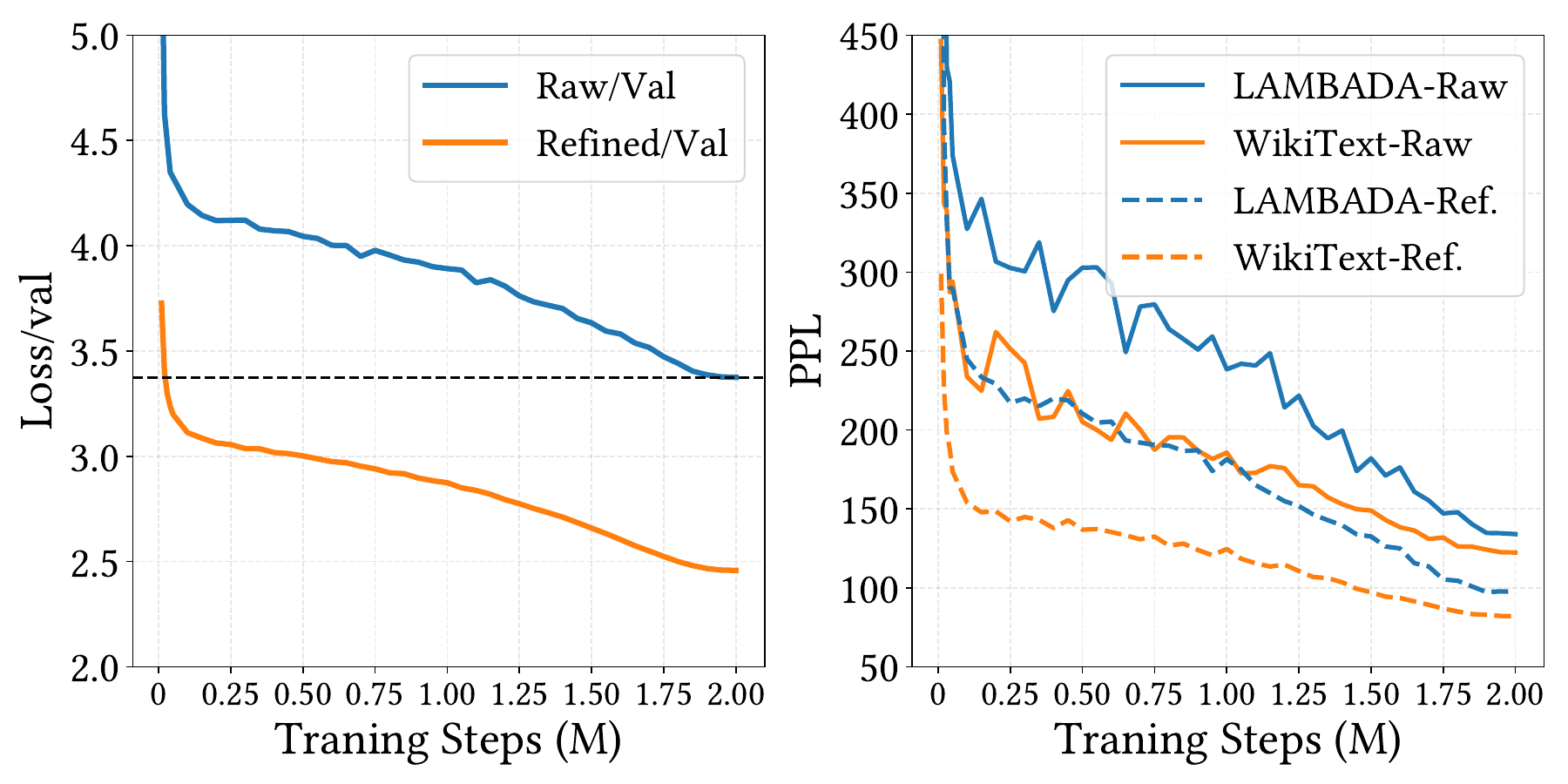}
  \setlength{\abovecaptionskip}{0.2cm}
  \setlength{\belowcaptionskip}{-0.75cm}
  \caption{Performance as a function of training steps. The left illustrates the decreasing loss on the validation set, while the right indicates trends of PPL on LAMBADA and WikiText103.}
  \label{fig:evaluation}
\end{figure}

\section*{ACKNOWLEDGEMENTS}
This research was supported by the Natural Science Foundation of China (61902209, 62377044, U2001212), and Beijing Outstanding Young Scientist Program (NO. BJJWZYJH012019100020098), Intelligent Social Governance Platform, Major Innovation \& Planning Interdisciplinary Platform for the "Double-First Class" Initiative, Renmin University of China and Public Computing Cloud,Renmin University of China.

\bibliographystyle{ACM-Reference-Format}
\bibliography{reference}

\end{document}